\documentclass{article} 
\usepackage{iclr2027_conference,times}

\usepackage{amsmath,amsfonts,bm}

\def\eqref#1{equation~\ref{#1}}

\def\1{\bm{1}}

\DeclareMathAlphabet{\mathsfit}{\encodingdefault}{\sfdefault}{m}{sl}
\SetMathAlphabet{\mathsfit}{bold}{\encodingdefault}{\sfdefault}{bx}{n}

\usepackage{graphicx}
\usepackage{booktabs}
\usepackage{float}
\usepackage{hyperref}
\usepackage{url}

\title{Shaping Latent Mixture-of-Experts: From Expert Geometry to Layerwise Capacity Allocation}

\author{\parbox[t]{\dimexpr\textwidth-2\tabcolsep\relax}{\centering
  Yaoxiang~Wang$^{1,2}$, Kai~Zhang$^{2}$, Ruobing~Xie$^{2}$, Lanrui~Wang$^{2}$,\\
  Guoliang~Zhao$^{2}$, Junhui~Li$^{3}$, Jinsong~Su$^{1}$\\[5pt]
  \normalfont $^{1}$Xiamen University \quad $^{2}$Tencent Hunyuan \quad $^{3}$Soochow University}}
  
\iclrfinalcopy 
\begin{document}

\maketitle
\lhead{}\renewcommand{\headrulewidth}{0pt}

\begin{abstract}
  Fine-grained mixture-of-experts (MoE) models commonly increase expert count by
  narrowing each expert, treating expert shape as a by-product of routing
  granularity. LatentMoE moves expert computation into a lower-dimensional
  space, exposing the input dimension and intermediate width of each expert as
  explicit design choices. We show that an expert's local Jacobian rank is
  jointly bounded by these two dimensions, and that balancing them maximizes the
  rank ceiling at fixed per-expert parameters. Controlled shape sweeps across
  model scales and routing granularities consistently favor near-square
  experts. We further use layerwise intrinsic dimension (ID) to guide capacity
  allocation across depth, co-varying the number and shape of experts while
  keeping per-layer parameter count, activated FLOPs, and communication volume
  constant. Compared with
  layer-homogeneous baselines, our ID-guided allocation achieves lower
  validation loss and stronger downstream performance. These results establish
  expert shape and layerwise allocation as practical design axes for MoEs.
\end{abstract}

\section{Introduction}

\begin{figure}[htb]
\centering
\includegraphics[width=0.92\linewidth]{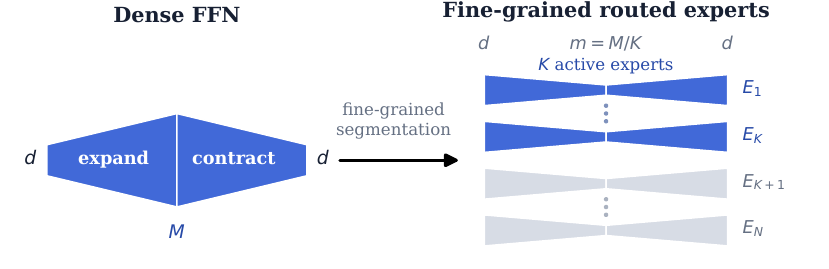}
\caption{From a wide dense FFN to fine-grained routed experts. Blue denotes
the dense FFN and the experts activated for one token; gray denotes inactive
experts. Let $M$ denote the dense FFN intermediate width and $K$ the number
of experts selected per token. Activated-width matching gives each selected
expert intermediate width $m=M/K$, making each expert much narrower.}
\label{fig:dense-to-fine-grained}
\end{figure}

Sparse mixture-of-experts (MoE) models have become a primary route to scaling
model capacity without paying the computational cost of activating every
parameter for every token
\citep{shazeer2017outrageously,lepikhin2021gshard,fedus2022switch,jiang2024mixtral}.
In a dense Transformer, the feed-forward network is intentionally wider than the residual
stream, reflecting a block-level allocation of parameters between attention
and feed-forward computation \citep{vaswani2017attention}. 
Recent MoE LLMs commonly replace the dense FFN with fine-grained experts while
matching the sum of the activated experts' widths to the dense FFN width
\citep{dai2024deepseekmoe,deepseekai2024v2,deepseekai2024v3}, as illustrated
in Figure~\ref{fig:dense-to-fine-grained}.

\begin{figure}[ht]
  \centering
  \includegraphics[width=0.95\linewidth]{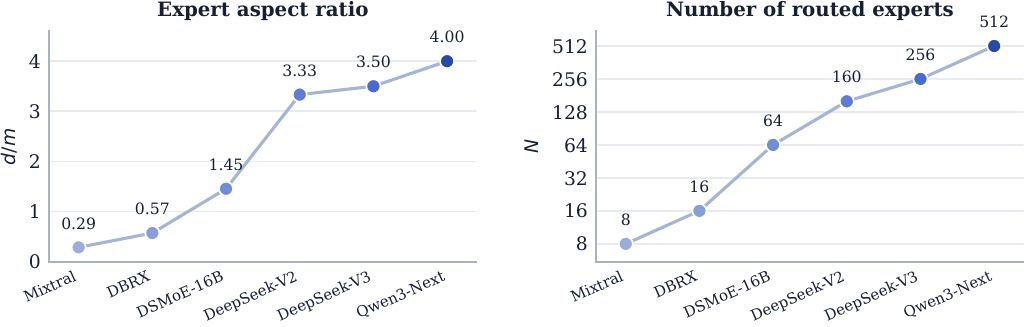}
  \caption{Fine-graining trends across representative MoE architectures.
  \textbf{Left:} the increasing per-expert aspect ratios $d/m$.
  \textbf{Right:} the increasing numbers of routed experts $N$ on a log scale. Values
  are taken from the corresponding model specifications
  \citep{jiang2024mixtral,databricks2024dbrx,dai2024deepseekmoe,deepseekai2024v2,deepseekai2024v3,qwen3next2025}.}
  \label{fig:expert-shape-trend}
\end{figure}

Once the block-level budget of attention and the FFN is chosen, the
attention configuration fixes the model hidden size $d$ (the width of the
residual stream), and every expert in a standard MoE inherits this input
dimension. Increasing the number of experts
under a matched FFN budget must therefore divide the budget into smaller
per-expert intermediate dimensions $m$, making the aspect ratio $d/m$ a
consequence of expert granularity rather than an independent design choice.
As Figure~\ref{fig:expert-shape-trend} shows, representative architectures
with more experts consequently exhibit increasingly large $d/m$ ratios, a
trend that may create a hidden bottleneck during training. Matching $Km$ to
the dense intermediate width preserves the activated parameter and compute
budget, but does not preserve how intermediate features are used. A dense FFN
applies the same fixed set of channels to every token, whereas an MoE router
selects different combinations of experts. Each expert must
therefore remain useful across many routing combinations and retain
independent representational capacity. Reducing $m$ may constrain
what each expert can represent even when the aggregate activated width $Km$
remains unchanged.

LatentMoE~\citep{elango2026latentmoe} reduces routed computation and
communication by applying experts in a lower-dimensional latent space. This
design was introduced for efficiency, but it
decouples the model hidden dimension from the expert input dimension and thus
exposes expert shape as a design choice. Since its original formulation fixes
a four-fold compression, we ask how to jointly choose the expert input and
intermediate dimensions under a fixed per-expert parameter budget.
Our expert-shape rationale starts from a simple local property: an
expert's Jacobian rank is bounded by both its input dimension and its
intermediate width. Under a fixed parameter budget, this ceiling is largest
when the two dimensions are balanced.
We test this implication in controlled shape sweeps and observe a U-shaped
validation-loss curve over aspect ratio, with the best performance in a
near-square region.

Within the near-square region, we further explore the effect of varying
per-expert capacity across depth. Intrinsic dimension
(ID), which estimates the number of locally independent degrees of freedom
occupied by a layer's input representations, varies substantially with
Transformer depth
\citep{valeriani2023geometry,joshi2025decisiongeometry}. We therefore hypothesize that
higher-ID layers benefit from greater per-expert capacity, whereas
repeating one expert shape at every layer may allocate capacity inefficiently.
We use the ID ordering to determine the direction of a coarse allocation, with
the expert dimensions selected under matched-budget and hardware constraints.
Prior heterogeneous designs vary expert counts or feed-forward widths across
layers \citep{jawahar2023automoe}, but doing so naively creates pronounced
per-layer compute and communication imbalance. This imbalance poses severe
systems-optimization challenges for pipeline partitioning and efficient MoE
kernels, making such designs difficult to deploy efficiently at scale. 
We
instead co-vary the number and shape of experts across depth to match
per-expert capacity to layerwise ID while keeping per-layer parameter count,
activated FLOPs, and communication volume nearly constant. The resulting
allocation is heterogeneous in capacity but nearly homogeneous from the
system's perspective. Our experiments show that this allocation improves
validation loss and downstream performance; ablations on uniform endpoints,
allocation direction, and regime boundaries further support its effectiveness
and robustness.

Our contributions are summarized as follows:
\begin{itemize}
  \item We establish expert shape as an overlooked but important MoE design
  variable and empirically validate the advantage of near-square experts,
  providing a low-cost rule for choosing the LatentMoE compression dimension.
  \item Under the near-square expert principle and guided by layerwise
  intrinsic dimension, we extend expert shaping from within-layer design to
  across-layer capacity allocation. This perspective challenges the usual
  assumption that every MoE layer should use the same expert configuration.
  \item We introduce an infrastructure-friendly layerwise allocation that
  keeps per-layer computation and communication nearly unchanged while
  achieving lower validation loss and stronger downstream-task performance.
\end{itemize}


\section{Preliminaries}
\label{sec:preliminaries}

Let $\bm{x}\in\mathbb{R}^{d}$ be the input to an MoE layer. A router selects
$K$ experts from a pool of $N$, denoted by
$\mathcal{T}_{K}(\bm{x})$, and assigns weights $p_i$. The routed path of a
standard MoE is
\[
  \bm{y}_{\mathrm{std}}
  =\sum_{i\in\mathcal{T}_{K}(\bm{x})}p_iE_i(\bm{x}),
  \qquad
  E_i:\mathbb{R}^{d}\rightarrow\mathbb{R}^{d},
\]
where each expert expands through an intermediate dimension $m$. Ignoring
architecture-dependent constants, the routed experts contain parameters
proportional to $Ndm$. For each token, activated expert computation is
proportional to $Kdm$, while the all-to-all communication volume for dispatch
and aggregation is proportional to $Kd$. The residual width $d$ therefore
determines both the expert input dimension and the width of the communicated
representations.

LatentMoE introduces learned projections
$W_{\downarrow}\in\mathbb{R}^{\ell\times d}$ and
$W_{\uparrow}\in\mathbb{R}^{d\times\ell}$ \citep{elango2026latentmoe}:
\[
  \bm{z}=W_{\downarrow}\bm{x}, \qquad
  \bm{y}_{\mathrm{latent}}
  =W_{\uparrow}\sum_{i\in\mathcal{T}_{K}(\bm{x})}
  p_i E_i(\bm{z}),
  \qquad
  E_i:\mathbb{R}^{\ell}\rightarrow\mathbb{R}^{\ell}.
\]
The routed expert path, including all-to-all dispatch and aggregation, now
operates in the $\ell$-dimensional space. Under the same accounting, routed
expert parameters, per-token activated computation, and per-token
communication volume are proportional to $N\ell m$, $K\ell m$, and $K\ell$,
respectively. Appendix~\ref{app:cost-accounting} provides the full accounting.

Let $\alpha=d/\ell$ denote the compression factor. The accuracy-oriented
LatentMoE variant keeps $m$ fixed and reinvests the compression savings by
setting $N'=\alpha N$ and $K'=\alpha K$. It then satisfies
\[
  N'\ell m=Ndm,\qquad
  K'\ell m=Kdm,\qquad
  K'\ell=Kd.
\]
Thus, under this routed-path accounting, it provides $\alpha$ times as many
candidate and activated experts without changing routed-expert parameters,
activated expert computation, or all-to-all communication volume. The
original study selected $\alpha=4$ empirically
\citep{elango2026latentmoe}, but did not provide a general principle for
jointly choosing the latent dimension $\ell$ and expert intermediate
dimension $m$. LatentMoE variants have been adopted by frontier models
including MAI-Thinking-1 and Kimi K3
\citep{microsoftai2026mai,kimiteam2026k3}, demonstrating their practical
relevance at large scale.

The key structural consequence is that LatentMoE decouples the expert input
dimension $\ell$ from the residual width $d$, making expert shape a
configurable design variable. We define the \emph{expert shape} as
$\ell\times m$. Since the per-expert parameter count is proportional to
$\ell m$, fixing this product allows the aspect ratio $\ell/m$ to vary without
changing the parameter budget. We ask which shape is most effective under
this constraint, first when shared across layers and then when allocated
layerwise to match the intrinsic dimension of the inputs.

\section{Expert Shaping}
\label{sec:expert-shaping}

\subsection{A Per-Expert Local-Rank Ceiling}

Consider a bias-free gated FFN expert
\[
E_i(\bm{z}) =
W_{o,i}\!\left[
  \phi(W_{g,i}\bm{z})\odot W_{u,i}\bm{z}
\right],
\]
where $W_{g,i},W_{u,i}\in\mathbb{R}^{m\times\ell}$ and
$W_{o,i}\in\mathbb{R}^{\ell\times m}$. This form includes the SwiGLU
experts used in our models \citep{shazeer2020glu}. Wherever $\phi$ is differentiable, its Jacobian
can be written as
\[
J_{E_i}(\bm{z}) =
W_{o,i}\!\left[
  D_{u,i}(\bm{z})W_{g,i}
  {}+{}D_{g,i}(\bm{z})W_{u,i}
\right],
\]
where $D_{u,i}$ and $D_{g,i}$ are diagonal $m\times m$ matrices determined
by the activations. The bracketed term has shape $m\times\ell$, and therefore
\[
  \operatorname{rank}J_{E_i}(\bm{z})\leq\min(\ell,m).
\]
The nonlinearity changes the entries of the local linearization but cannot
bypass the $m$-dimensional intermediate representation.
We denote this ceiling by $q=\min(\ell,m)$. At a fixed per-expert parameter
budget, $\ell m$ is fixed, so $q$ is maximized when $\ell=m$. The bound motivates the hypothesis that strongly
unbalanced shapes will underperform near-square ones when per-expert capacity
is limiting.

Across our shape sweeps, expert weights and routed-token outputs are
essentially full rank, and trained local Jacobians nearly attain $q$. This
supports $q$ as an empirically relevant capacity proxy;
Appendix~\ref{app:realized-rank} reports the estimators, controls, and full
results.

\subsection{Controlled Shape Sweeps}

We test this hypothesis with sparse LatentMoEs at 3.4B and 8B scales. At each
scale, we sweep $\ell/m$ under both $(N,K)=(256,8)$ and $(512,16)$; each
token therefore activates only 3.1\% of the routed experts, reflecting the
very sparse, fine-grained regime used by frontier MoEs. Within each sweep,
$N$ and $K$ are fixed, $\ell m$ is held constant, and all
other training choices are shared. Since changing $\ell$ also changes the projection parameters, we
adjust the full-width shared-expert dimension $m_s$ to keep the total
per-layer parameter count nearly constant. Appendix~\ref{app:training-details} gives the
training details. Appendix~\ref{app:fixed-shared-control} reports a
complementary fixed-$m_s$ control that preserves the same qualitative result
without this parameter-matching adjustment.

\begin{figure}[htb]
\centering
\includegraphics[width=0.98\linewidth]{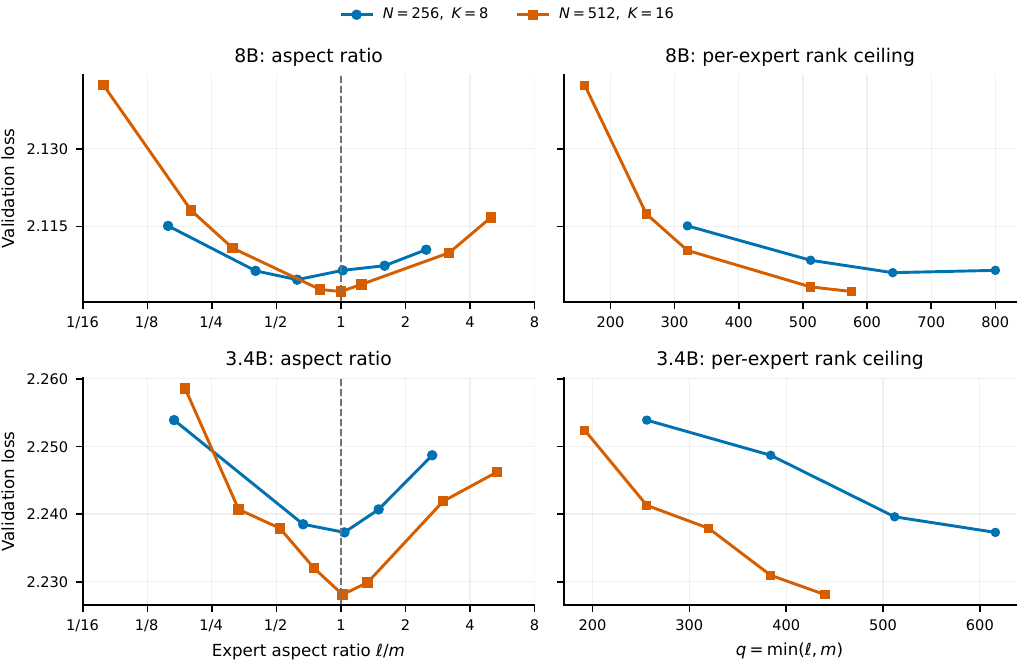}
\caption{Controlled expert-shape sweeps at two model scales and two routing
granularities. \textbf{Left:} validation loss versus aspect ratio; the dashed
line marks a square expert. \textbf{Right:} the same runs plotted against the
per-expert local-rank ceiling $q=\min(\ell,m)$, with loss averaged over shapes
sharing the same $q$. Lower is better; Appendix~\ref{app:shape-sweep-results}
gives all configurations and losses.}
\label{fig:shape-sweeps}
\end{figure}

Figure~\ref{fig:shape-sweeps} shows the same U-shaped dependence on aspect
ratio at both scales and routing granularities: highly unbalanced experts
deteriorate in either direction, while the best region lies near a square.
Operationally, the tested shapes with $\lvert \ell/m-1\rvert\leq 0.4$ lie
in or near the best-performing plateau across these sweeps, providing a
practical tolerance around the idealized square shape.
Across the parameter-matched sweeps, the resulting validation-loss gaps are
substantial relative to empirical run-to-run variation; Appendix
Section~\ref{app:training-details} gives the supporting replication details.
Under fixed per-expert parameters, validation loss broadly improves as the
local-rank ceiling $q$ grows and then plateaus.
They yield a simple design rule: begin with a hardware-feasible
shape near $\ell=m$, then tune within the resulting plateau if needed.

\section{Layerwise ID-Guided Capacity Allocation}
\label{sec:layerwise-id-allocation}

Section~\ref{sec:expert-shaping} holds the expert budget fixed and optimizes
its shape. We next ask whether the per-expert budget itself should be
uniform across depth. LatentMoE exposes a trade-off: compressing token
representations and the per-expert parameter budget can hurt
performance by constraining the capacity of each individual expert, while
reinvesting the savings into more experts can improve performance.
A uniform compression implicitly assumes that
this trade-off is the same at every layer. However, prior work shows that the
geometry and intrinsic dimension (ID) of token representations vary systematically
with depth \citep{valeriani2023geometry,joshi2025decisiongeometry}, suggesting
that different layers may tolerate different degrees of representation and
per-expert compression. We therefore use
$q_t=\min(\ell_t,m_t)$ as a local-capacity proxy for layer $t$ and test the
directional hypothesis that higher-ID regions may benefit from a larger
$q_t$, whereas lower-ID regions may tolerate a smaller one. Because ID and
$q_t$ measure different quantities, the allocation uses their ordinal
relation as a directional signal.

\subsection{Estimating Layerwise Intrinsic Dimension}

Intrinsic dimension estimates the number of local degrees of freedom
occupied by representations, which may be much smaller than their ambient
width \citep{ansuini2019intrinsic}. Prior work has found that this quantity varies substantially with
Transformer depth
\citep{valeriani2023geometry,joshi2025decisiongeometry}. We measure it using TwoNN
\citep{facco2017twonn}.

Given representation samples
$\{\bm{h}^{(t)}_j\}$ at a block boundary, let $r_{j,1}$ and $r_{j,2}$ be the
distances from sample $j$ to its first and second nearest neighbors, and set
$\mu_j=r_{j,2}/r_{j,1}$. Under locally constant sampling density on an
$r_t$-dimensional manifold,
\[
  F(\mu)=1-\mu^{-r_t},
  \qquad
  -\log(1-F(\mu))=r_t\log\mu.
\]
We estimate $r_t$ by fitting this slope through the origin, discarding the
largest 10\% of neighbor ratios for robustness.

\begin{figure}[htbp]
\centering
\includegraphics[width=0.98\linewidth]{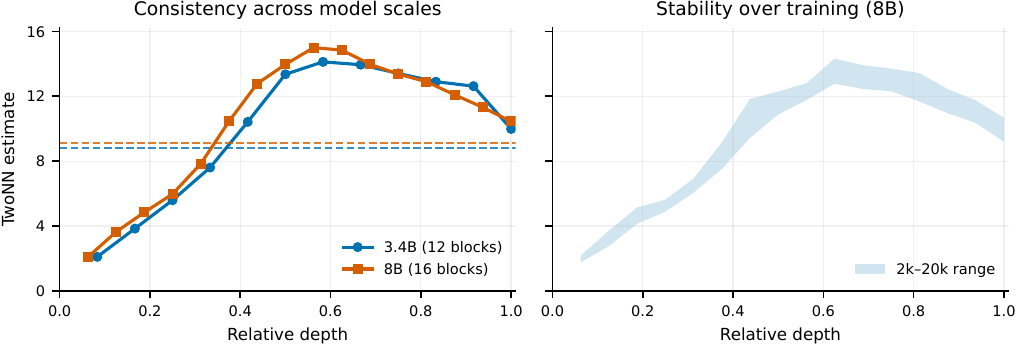}
\caption{Layerwise TwoNN profiles. \textbf{Left:} final profiles at two model
scales exhibit closely aligned variation over relative depth; the
color-matched horizontal dashed lines mark the thresholds used to separate
the shallow and deeper regimes.
\textbf{Right:} the pointwise range across a separate 8B checkpoint series
shows that the depth-wise profile is already visible early and persists
through training.}
\label{fig:id-profiles}
\end{figure}
\vspace{-4pt}

Figure~\ref{fig:id-profiles} shows a consistent qualitative pattern: ID rises
through the shallow layers, peaks in the middle-to-deep portion of the model,
and declines gradually near the output. The same depth-wise ordering is
visible across model scales and from an early training checkpoint onward.
Consequently, the allocation can be chosen from a short proxy run rather than
requiring a fully trained base model.
Appendix~\ref{app:id-details} gives the estimator, sampling, and checkpoint
details.

\subsection{Constant-Cost Capacity Allocation}

Given a target ordering of per-expert capacities, a naive heterogeneous model changes FFN width or expert count alone and
thereby creates layerwise compute and communication imbalance. LatentMoE
provides enough degrees of freedom to vary per-expert capacity while avoiding
this imbalance. Up to constants shared by all layers, define the routed
computation, routed-expert parameters, and one-way dispatch volume as
\[
  C_t=K_t\ell_t m_t,\qquad
  P_t=N_t\ell_t m_t,\qquad
  V_t=K_t\ell_t.
\]
Requiring $C_t=C^\star$, $P_t=P^\star$, and $V_t=V^\star$ implies
\[
  m_t=\frac{C^\star}{V^\star},\qquad
  \frac{N_t}{K_t}=\frac{P^\star}{C^\star},\qquad
  K_t=\frac{V^\star}{\ell_t}.
\]
Thus $m_t$ remains approximately constant, the sparsity ratio $N_t/K_t$
is preserved, and increasing $\ell_t$ is compensated by decreasing both
$K_t$ and $N_t$. This can raise the per-expert ceiling $q_t$ without changing
the aggregate routed budget.

We additionally keep
\[
  Q_t=2\ell_t+N_t+3m_{s,t}
\]
nearly constant. Multiplication by $d$ accounts for the two latent
projections, router, and full-width shared expert, so matching $Q_t$ controls
the remaining dominant MoE parameters and computation. Appendix~\ref{app:cost-accounting} expands this accounting, with particular
attention to expert-parallel communication. Finally, we use near-square rather
than exactly square experts, allowing up to a 30\% aspect-ratio deviation to
align per-layer compute and communication while satisfying grouped-GEMM and
expert-parallel divisibility requirements. The equalities are approximate
where integer hardware constraints require rounding. The construction is
therefore heterogeneous in per-expert capacity but nearly homogeneous in
per-layer parameter count, activated FLOPs, and communication volume.

\paragraph{Practical two-segment allocation.}
Although a future calibrated method could assign a distinct configuration to
every layer, our experiments deliberately use only two regimes. For a simple
implementation with minimal kernel and configuration complexity,
we partition the MoE blocks at a clear separation in the measured ID profile:
the largest ID in the shallow prefix remains below the smallest ID in the
remaining blocks. 
In our profiles, the latter minimum typically occurs at the
final block given the late-depth decline. We assign the lower-capacity
configuration to the shallow prefix and the higher-capacity configuration to
the remaining blocks. This data-dependent split avoids a scale-specific
numerical threshold and is an implementation choice rather than a restriction
of the allocation rule. Table~\ref{tab:layerwise-configs} lists the resulting
configurations.

\section{Experiments}
\label{sec:experiments}

We separate end-task evaluation from controlled diagnosis. We first compare
complete models on downstream benchmarks in Section~\ref{sec:downstream-results}.
We then use validation loss in Section~\ref{sec:pretraining-results} to isolate
the effects of expert shape, allocation direction, and the regime boundary.
This distinction is important because the expected differences between
pretraining architectures are small relative to the resolution of individual
downstream tasks.

\subsection{Experimental Setup}
\label{sec:experimental-setup}

\paragraph{Pretraining.}
The controlled shape sweeps and layerwise experiments use the models and
training protocols detailed in Appendix~\ref{app:training-details}. For the
layerwise experiments, we use the configurations in
Table~\ref{tab:layerwise-configs} and compare their ID-guided placement with
uniform and exchanged controls. Within each comparison, the data, optimizer,
and training schedule are unchanged.

\paragraph{Compared models.}
Table~\ref{tab:layerwise-configs} lists the practical configurations resulting
from the two-segment allocation in Section~\ref{sec:layerwise-id-allocation}. 
At 3.4B, routed computation and routed parameters match exactly between
regimes; dispatch differs by 1.8\%. At 8B,
routed computation, routed parameters, and dispatch each differ by at most
0.32\%. We evaluate four settings at each scale, matching their total and activated
parameter budgets within that scale. The base model's experts
receive the uncompressed residual-stream input, with width 1,280 at 8B and
1,024 at 3.4B. The two uniform LatentMoE models use compression factors
$\alpha=2$ and $\alpha=4$, scaling $(N,K)$ to $(512,16)$ and $(1024,32)$,
respectively, following the original LatentMoE scaling rule
\citep{elango2026latentmoe}. Note that the
$\alpha=2$ baseline is already near-square. The layerwise model uses the two-regime
allocation in Table~\ref{tab:layerwise-configs}. Appendix
Table~\ref{tab:downstream-models} gives the full configurations.

\begin{table}[H]
\vspace{-4pt}
\centering
\small
\setlength{\tabcolsep}{8pt}
\caption{Two-segment ID-guided configurations used in the experiments.
Block 0 is dense; the listed indices are MoE blocks. Here $m_s$ is the
full-width shared-expert intermediate dimension.}
\label{tab:layerwise-configs}
\begin{tabular}{llrrrrr}
\toprule
Scale & Regime (blocks) & $\ell$ & $m$ & $N$ & $K$ & $m_s$ \\
\midrule
3.4B & shallow (1--4)  & 352 & 448 & 640 & 20 & 490  \\
     & deeper  (5--11) & 512 & 440 & 448 & 14 & 448  \\
8B   & shallow (1--5)  & 464 & 512 & 704 & 22 & 1077 \\
     & deeper  (6--15) & 640 & 512 & 512 & 16 & 1024 \\
\bottomrule
\end{tabular}
\end{table}
\vspace{-4pt}

\paragraph{Downstream protocol.}
We use the full OLMES \texttt{core\_9mcqa} suite \citep{gu2024olmes}:
ARC-Easy, ARC-Challenge, BoolQ, CommonsenseQA, HellaSwag, OpenBookQA, PIQA,
SocialIQA, and WinoGrande. We follow the standard OLMES-v0.1 5-shot protocol
and report the \texttt{mc\_or\_rc} score, which selects the better of the
multiple-choice and ranked-classification formats for each model--task pair.

\subsection{Main Downstream Results}
\label{sec:downstream-results}
\begin{table}[H]
\vspace{-4pt}
\centering

\caption{Task-level downstream results using the \texttt{mc\_or\_rc} score
(\%). Task columns and Avg.\ report means over three training runs; Avg.\ additionally
reports the standard deviation across runs.}
\label{tab:downstream-main}
\resizebox{\textwidth}{!}{%
\begin{tabular}{l*{10}{r}}
\toprule
Setting & ARC-E & ARC-C & BoolQ & CSQA & Hella. & OBQA
        & PIQA & SIQA & Wino. & Avg. $\pm$ s.d. \\
\midrule
\multicolumn{11}{c}{\textbf{3.4B}} \\
\midrule
Base
  & 60.34 & 33.79 & 55.66 & 50.21 & 45.13 & 40.13
  & 69.08 & 45.21 & 52.27 & $50.20\pm0.15$ \\
LatentMoE $\alpha=2$
  & 61.64 & 34.81 & 59.78 & 51.49 & 45.90 & 41.47
  & 69.88 & 45.58 & 51.93 & $51.39\pm0.16$ \\
LatentMoE $\alpha=4$
  & 61.93 & 33.93 & 61.11 & 51.00 & 45.62 & 39.80
  & 69.37 & 45.47 & 51.38 & $51.07\pm0.24$ \\
Ours
  & 62.18 & 36.01 & 60.57 & 52.14 & 46.86 & 42.47
  & 70.66 & 45.17 & 52.16 & $\mathbf{52.02\pm0.30}$ \\
\midrule
\multicolumn{11}{c}{\textbf{8B}} \\
\midrule
Base
  & 65.67 & 40.10 & 57.19 & 56.79 & 54.64 & 40.73
  & 71.54 & 50.02 & 56.96 & $54.85\pm0.43$ \\
LatentMoE $\alpha=2$
  & 66.50 & 39.33 & 63.56 & 56.84 & 55.13 & 44.53
  & 72.91 & 49.38 & 55.43 & $55.96\pm0.21$ \\
LatentMoE $\alpha=4$
  & 66.36 & 41.84 & 61.34 & 57.82 & 54.43 & 44.67
  & 71.73 & 48.11 & 54.41 & $55.63\pm0.20$ \\
Ours
  & 68.84 & 41.81 & 60.49 & 60.06 & 56.37 & 47.00
  & 72.34 & 49.78 & 55.12 & $\mathbf{56.87\pm0.37}$ \\
\bottomrule
\end{tabular}
}
\end{table}

At both scales, the ID-guided layerwise allocation achieves the highest mean
score across three runs, outperforming both uniform LatentMoE baselines and
the base model. This provides a descriptive downstream complement to the
matched validation-loss evidence, with the reported standard deviations
summarizing run-level variation in the aggregate score.

\subsection{Analysis}
\label{sec:pretraining-results}

Downstream accuracy is too coarse for the controlled fine-grained ablations.
We therefore use validation loss exclusively in this subsection.

\paragraph{Shape must be optimized before granularity.}
Across the four sweeps in Figure~\ref{fig:shape-sweeps}, moving from an
extreme aspect ratio toward the near-square region consistently reduces
validation loss. The trend appears at both routing granularities and model
scales, and survives the fixed-$m_s$ control in
Appendix~\ref{app:fixed-shared-control}. These comparisons isolate expert shape
from the number of available and activated experts: routing granularity alone
leaves the shape-dependent per-expert constraint in place.

\paragraph{ID-guided placement beats both matched endpoints.}
Table~\ref{tab:layerwise-val} compares the layerwise allocation with the two
uniform models obtained by repeating either of its regimes at every MoE block.
The uniform narrow model applies a lower per-expert rank ceiling even to the
higher-ID later blocks, where it can become restrictive. The uniform wide
model raises this ceiling throughout but also assigns the same capacity to
the lower-ID shallow blocks, where the additional per-expert capacity is less
useful. The ID-guided model instead places the two regimes according to the
observed ordinal profile and improves over both uniform endpoints at both
scales. This establishes a benefit beyond interpolation between a strong and
a weak uniform model. Since both uniform endpoints are themselves near-square
($\lvert\ell/m-1\rvert\leq0.4$), this comparison holds expert shape fixed and
isolates the across-depth allocation as the source of the gain. Broader
comparisons among depth-correlated allocation signals can further refine the
choice of guidance signal.

\begin{table}[H]
\centering
\small
\setlength{\tabcolsep}{2.5pt}
\caption{Configurations and validation loss for matched layerwise controls.
Narrow and wide denote the lower- and higher-capacity regimes in
Table~\ref{tab:layerwise-configs}. The fixed-split exchange preserves the
ID-guided boundary while swapping regimes; the matched-count exchange
reverses their order while preserving how many blocks use each regime.
Allocations are ordered from shallow to deep. Lower loss is better.}
\label{tab:layerwise-val}
\begin{tabular}{lccr}
\toprule
\multicolumn{4}{c}{\textbf{3.4B} (11 MoE blocks)} \\
\midrule
Setting & $(\ell,m)$ & $(K,N)$ & Val.\ loss \\
\midrule
Uniform narrow
  & $11\times(352,448)$
  & $11\times(20,640)$
  & 2.2391 \\
Uniform wide
  & $11\times(512,440)$
  & $11\times(14,448)$
  & 2.2305 \\
ID-guided
  & $4\times(352,448)+7\times(512,440)$
  & $4\times(20,640)+7\times(14,448)$
  & $\mathbf{2.2234}$ \\
Exchange (fixed split)
  & $4\times(512,440)+7\times(352,448)$
  & $4\times(14,448)+7\times(20,640)$
  & 2.2422 \\
Exchange (matched count)
  & $7\times(512,440)+4\times(352,448)$
  & $7\times(14,448)+4\times(20,640)$
  & 2.2433 \\
\midrule
\multicolumn{4}{c}{\textbf{8B} (15 MoE blocks)} \\
\midrule
Setting & $(\ell,m)$ & $(K,N)$ & Val.\ loss \\
\midrule
Uniform narrow
  & $15\times(464,512)$
  & $15\times(22,704)$
  & 2.1041 \\
Uniform wide
  & $15\times(640,512)$
  & $15\times(16,512)$
  & 2.1038 \\
ID-guided
  & $5\times(464,512)+10\times(640,512)$
  & $5\times(22,704)+10\times(16,512)$
  & $\mathbf{2.0956}$ \\
Exchange (fixed split)
  & $5\times(640,512)+10\times(464,512)$
  & $5\times(16,512)+10\times(22,704)$
  & 2.1092 \\
Exchange (matched count)
  & $10\times(640,512)+5\times(464,512)$
  & $10\times(16,512)+5\times(22,704)$
  & 2.1088 \\
\bottomrule
\end{tabular}
\end{table}
\vspace{-4pt}

\paragraph{Direction, not heterogeneity alone, produces the gain.}
The two exchanged controls isolate complementary alternatives. The
fixed-split version preserves the ID-guided boundary but reverses which regime
appears on each side. The matched-count version preserves the exact multiset
of narrow and wide blocks and changes only their order. Both reversals worsen
validation loss at both scales and perform below both uniform endpoints.
In particular, the matched-count result rules out the possibility that the
gain comes merely from mixing the two regimes or from using more wide blocks;
within the tested controls, their shallow-to-deep ordering is important. This
result supports the allocation direction suggested by the ID profile.

\paragraph{The regime boundary is robust.}
Moving the shallow--deep boundary by one block in either direction changes
validation loss only slightly, and every tested boundary remains better than
the stronger uniform endpoint. Assigning the final block back to the shallow
configuration also degrades performance, suggesting that the late ID decline
does not warrant an additional regime switch. A coarse two-part allocation
therefore captures most of the benefit without fitting an exact boundary to
validation loss. Appendix~\ref{app:layerwise-ablation} reports the complete
configurations and results.

\begin{figure}[htbp]
\centering
\includegraphics[width=0.98\linewidth]{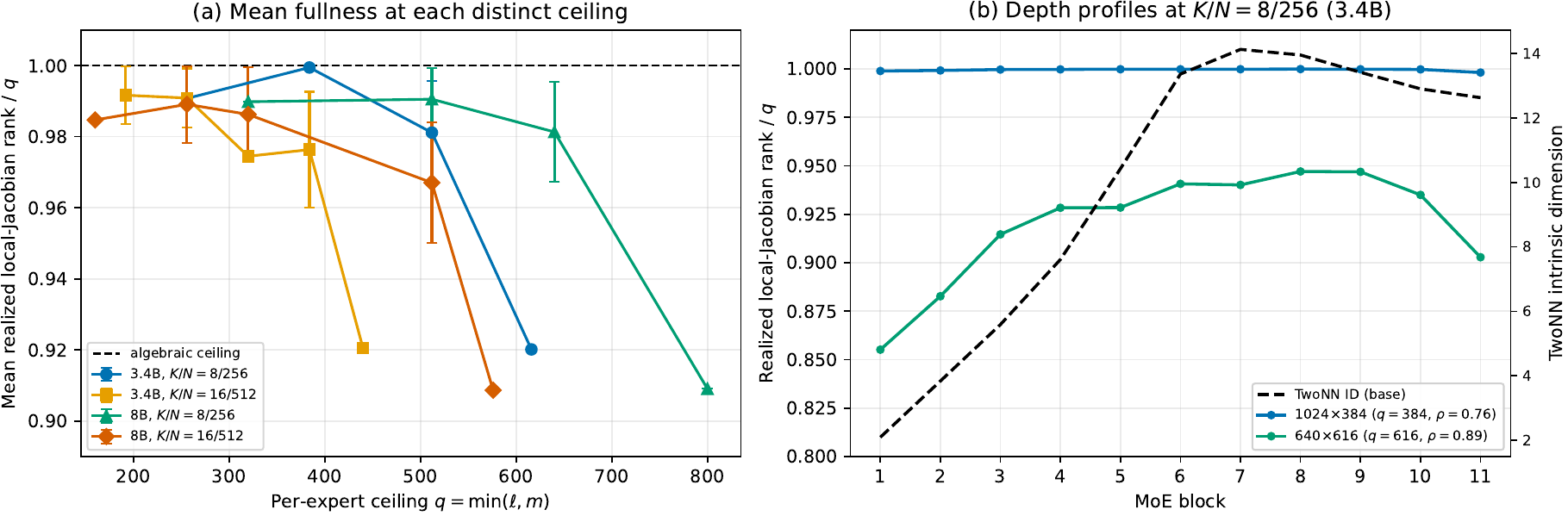}
\caption{Realized local-Jacobian rank. \textbf{Left:} depth-averaged rank
normalized by $q$; markers show means, bars show ranges, and the dashed line
is the ceiling. \textbf{Right:} blockwise normalized rank for two fixed-$q$
3.4B models; the dashed curve is base-model TwoNN ID.}
\label{fig:rank-diagnostics}
\end{figure}

\paragraph{Scope of the rank proxy.}
When the four shape sweeps are grouped by
$q=\min(\ell,m)$, mean validation loss decreases as $q$ grows and then
plateaus, consistent with reducing a per-expert capacity constraint. However,
transposed shapes with the same $q$ are not always equivalent. The fixed-$m_s$
experiment confirms that the broad near-square advantage remains after
removing shared-width reallocation, but it also shows that $q$ is not a
complete performance model. Figure~\ref{fig:rank-diagnostics} provides two
complementary diagnostics. First, local Jacobians closely approach $q$ across
the full shape sweep, showing that the single-expert ceiling is attained rather
than vacuous. At the mixture level, the selected expert Jacobians can
collectively span the latent space, making $q$ specifically a per-expert
capacity proxy. Second,
uniform-shape models with fixed $q$ exhibit depth-wise variation in realized
rank aligned with the base-model ID profile, showing that the available
ceiling and its realized use are distinct. Moreover, transposed shapes realize
nearly the same rank without reaching the same loss. Shape orientation
($\ell<m$ versus $\ell>m$), optimization, and interactions with the shared
path therefore remain relevant once the rank ceiling is no longer limiting.
Appendix~\ref{app:realized-rank} reports the estimator, controls, and complete
measurements.

\section{Related Work}
\label{sec:related-work}

Transformer capacity can be made nonuniform across layers, tokens, or experts.
AutoMoE searches over expert placement, per-layer expert count, and FFN width
under global FLOP and latency constraints \citep{jawahar2023automoe}.
Fixed-budget methods taper MLP width or derive layerwise FFN widths from
representation geometry \citep{bayat2026tapered,mudarisov2026geometryguided};
MAPLE reallocates routed experts by layer sensitivity \citep{li2026maple}, and
Prism schedules attention heads across depth \citep{aggarwal2026prism}.
Dynamic alternatives route selected tokens through blocks or vary token-level
expert activation \citep{raposo2024mixturedepths,gulmez2026dynamoe}. Together,
these studies establish nonuniform allocation as useful, but generally change
per-layer or per-token workloads, or target a component other than latent
expert shape.

Within a layer, heterogeneous-expert methods route tokens among different-width
experts and require mixed-size execution or explicit load balancing
\citep{wang2025hmoe,sun2024diversesize,ma2026mohge}. Our construction instead
uses one static expert shape and a fixed Top-$K$ rule per layer, co-varying
$(\ell,m,N,K)$ to keep routed computation, routed parameters, and dispatch
volume nearly constant. Its schedule follows the ordinal intrinsic-dimension
profile \citep{valeriani2023geometry,joshi2025decisiongeometry}, using input ID
to set the local-rank ceiling while retaining homogeneous grouped GEMMs and
predictable per-layer workloads.

\section{Conclusion}
\label{sec:conclusion}

We identify expert shape as a first-class design variable in latent MoEs. The
per-expert local-rank ceiling favors a near-square factorization, and controlled
sweeps consistently support this design rule. We further propose an
intrinsic-dimension-guided allocation that varies per-expert capacity across
depth while preserving aggregate compute and communication. Its practical
instantiation improves matched uniform controls, whereas reversing the
capacity placement removes the gain. The rank ceiling remains a single-expert proxy and intrinsic
dimension an ordinal allocation signal. Together, they provide practical
guidance for expert shaping and coarse layerwise allocation, while a
fine-grained mapping from representation geometry to per-layer capacity
remains a promising extension.

\clearpage

\subsection*{AI use statement}

In this work, we used generative AI tools for language polishing, LaTeX
formatting, and drafting and refining code. The authors
manually reviewed and edited all AI-assisted text, inspected and tested the
generated code, and checked the resulting figures. 
We did not use generative AI tools for any of the
following purposes: generating synthetic datasets; developing theoretical
models or conceptual frameworks; formulating mathematical claims; providing
critical ingredients for proving mathematical claims; assisting in the writing
of proofs; proposing or refining hypotheses; designing or providing feedback
on research methodology or experiments; implementing methods; assisting with
translation; cleaning or reformatting datasets; supporting qualitative or
thematic data analysis; or interpreting results. We take responsibility for
the final content of this work, including all text, claims, code, and artifacts
produced with the aid of generative AI.

\subsection*{Ethics statement}

This work uses existing language-model corpora and benchmarks and involves no
human-subject experiments or new personal-data collection. It inherits
potential bias, privacy, and misuse risks from the data and general-purpose
language models. Training also incurs computational and energy costs.

\subsection*{Reproducibility statement}

Sections~\ref{sec:preliminaries}--\ref{sec:experiments} define the architecture
and budget quantities, the shaping and allocation methods, and the evaluation
protocol. Appendix~\ref{app:training-details} reports the datasets,
optimization settings, and model configurations. The remaining appendices
provide complete shape sweeps, matched controls and ablations, parameter,
compute, and communication accounting, and details of the realized-rank and
intrinsic-dimension estimators.

\bibliography{iclr2027_conference}
\bibliographystyle{iclr2027_conference}

\clearpage
\appendix

\section{Additional Related Work}

Sparse MoEs scale parameter capacity by routing each token to a small subset
of experts \citep{shazeer2017outrageously}. GShard and Switch Transformer
established scalable expert parallelism and load-balancing strategies for
Transformer models \citep{lepikhin2021gshard,fedus2022switch}, and later
systems such as Mixtral and DBRX demonstrated the effectiveness of sparse MoEs
at large language-model scale \citep{jiang2024mixtral,databricks2024dbrx}.
DeepSeekMoE and its successors pursue finer expert segmentation by increasing
$N$ and $K$ while reducing individual expert width
\citep{dai2024deepseekmoe,deepseekai2024v2,deepseekai2024v3}. This accounting
couples routing granularity to increasingly narrow experts. LatentMoE instead
uses a projected space to decouple $N$ and $K$ from the residual width
\citep{elango2026latentmoe}, and appears in systems including MAI-Thinking-1
and Kimi K3 \citep{microsoftai2026mai,kimiteam2026k3}.
Related latent-factorization work compresses pretrained MoE experts into
shared projections and expert-specific low-dimensional transformations
\citep{liu2025molae}; it targets post-training parameter efficiency rather
than expert shape and routed communication during pretraining.

These developments primarily optimize routing, expert count, compression, or
aggregate budget. Our work is orthogonal: it isolates the two-dimensional
factorization $\ell\times m$, relates it to a per-expert local-rank ceiling,
and allocates that capacity across depth. Rank matching can therefore be
combined with existing routing and load-balancing methods.

\section{Training, Evaluation, and Systems Details}
\label{app:training-details}

\subsection{Pretraining Details}

The shape sweeps use a 3.4B-A250M model with hidden size 1,024 and 12
Transformer blocks, and an 8B-A560M model with hidden size 1,280 and 16
blocks. The first block is dense and all remaining blocks use sparse
LatentMoE. The backbone follows the DeepSeek-V3 architecture
\citep{deepseekai2024v3}, with its routed experts replaced by the LatentMoE
structure described in Section~\ref{sec:preliminaries}. Both scales use a
sequence length of 4,096 and the Adam optimizer with
$(\beta_1,\beta_2)=(0.9,0.95)$. The 3.4B-A250M model uses a learning rate of
$4\times10^{-4}$, a batch size of 2M tokens, and 1,500 warmup steps; the
8B-A560M model uses a learning rate of $3\times10^{-4}$, a batch size of 2M
tokens, and 2,000 warmup steps. The models train for 20B and 40B tokens,
respectively. We train on NVIDIA Nemotron-CC \citep{su2025nemotroncc}. Following
DeepSeek-V3, we combine an auxiliary load-balancing loss with coefficient
$10^{-4}$ and expert-bias updates with rate $10^{-3}$. Within a sweep, the
data mixture, initialization protocol, optimizer configuration, batch
construction, router, attention blocks, and expert activation are held fixed.
The layerwise experiments use the same scale-specific training setup; within
each set of controls, only the placement of the MoE configurations in
Table~\ref{tab:layerwise-configs} changes.

\paragraph{Run-to-run variation.}
We estimate the validation-loss noise scale from five independent training
runs of the base configuration under the same protocol. Their final
validation losses have a standard deviation of $1.2\times10^{-3}$ at
3.4B-A250M and $1.5\times10^{-3}$ at 8B-A560M. Because
repeating every point in the two-scale shape sweep would be computationally
prohibitive, each shape is trained once. The largest
within-sweep gaps, $1.04\times10^{-2}$--$3.99\times10^{-2}$, are roughly
7--27 times this empirical noise scale, making the observed separation
between extreme and near-square shapes substantially larger than ordinary
run-to-run variation.

\paragraph{Downstream model configurations.}
Table~\ref{tab:downstream-models} lists the four downstream settings at each
scale. Total and activated parameter budgets are matched within each scale;
internal checkpoint identifiers are omitted.

\begin{table}[H]
\centering
\small
\setlength{\tabcolsep}{4pt}
\caption{Models used in the downstream evaluation. For uniform models,
$\alpha=d/\ell$. Expert input denotes the representation presented to each
expert: the base receives the uncompressed residual stream, whereas LatentMoE
operates at latent width $\ell$. The layerwise entries use the two
configurations in Table~\ref{tab:layerwise-configs}.}
\label{tab:downstream-models}
\begin{tabular}{cllccc}
\toprule
Scale & Setting & Expert input & $m$ & $(N,K)$ & $m_s$ \\
\midrule
8B   & Base                  & uncompressed (1,280) & 512 & $(256,8)$   & 768 \\
     & LatentMoE, $\alpha=4$ & latent ($\ell=320$)  & 512 & $(1024,32)$ & 1067 \\
     & LatentMoE, $\alpha=2$ & latent ($\ell=640$)  & 512 & $(512,16)$  & 1109 \\
     & Ours: ID-guided layerwise
     & \multicolumn{4}{c}{two regimes in Table~\ref{tab:layerwise-configs}} \\
\midrule
3.4B & Base                  & uncompressed (1,024) & 384 & $(256,8)$   & 384 \\
     & LatentMoE, $\alpha=4$ & latent ($\ell=256$)  & 384 & $(1024,32)$ & 597 \\
     & LatentMoE, $\alpha=2$ & latent ($\ell=512$)  & 384 & $(512,16)$  & 427 \\
     & Ours: ID-guided layerwise
     & \multicolumn{4}{c}{two regimes in Table~\ref{tab:layerwise-configs}} \\
\bottomrule
\end{tabular}
\end{table}

\subsection{Parameter, Compute, and Communication Accounting}
\label{app:cost-accounting}

Section~\ref{sec:layerwise-id-allocation} defines the matched quantities up to
constants shared across layers. Here we expand those constants and distinguish
model parameters, activated computation, and expert-parallel communication.
We consider one LatentMoE layer; unchanged attention and backbone components
are outside this comparison. A
SwiGLU expert contains three weight matrices: two input projections and one
output projection.

\begin{figure}[H]
\centering
\includegraphics[width=\linewidth]{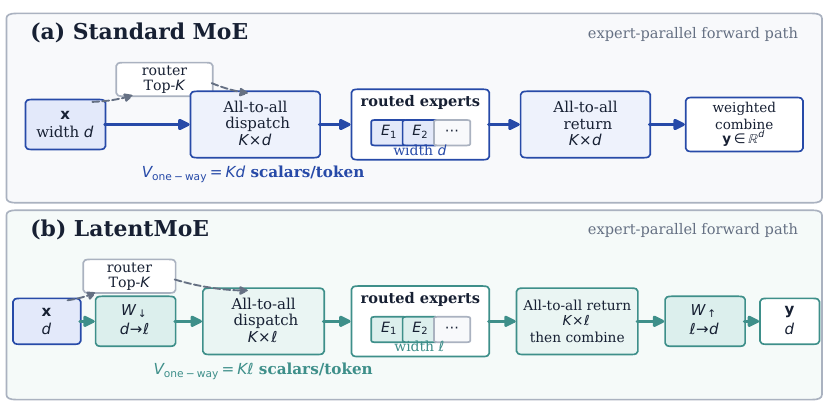}
\caption{Expert-parallel communication paths.
\textbf{Top:} a standard MoE dispatches and returns $K$ vectors of width $d$
per token. \textbf{Bottom:} LatentMoE retains routing decisions in the model
hidden space but places the shared down- and up-projections outside the two
all-to-all collectives, reducing each routed vector from width $d$ to
$\ell$. Gray dashed arrows show the local routing computation and Top-$K$
assignments; colored arrows show the activation path.}
\label{fig:communication-accounting}
\end{figure}

\paragraph{Parameter count.}
The $N_t$ routed experts contain $3N_t\ell_t m_t$ parameters. The latent
down- and up-projections contain $2d\ell_t$ parameters, the router contains
$dN_t$, and the full-width shared expert contains $3dm_{s,t}$. The
layer-dependent unique trainable parameters are therefore
\[
  \operatorname{Params}_t
  \approx
  3N_t\ell_t m_t
  +d\!\left(2\ell_t+N_t+3m_{s,t}\right)
  =3P_t+dQ_t.
\]
Thus, $P_t$ in the main text controls the routed-expert parameter pool, while
$Q_t$ controls the remaining dominant parameters.

\paragraph{Activated computation.}
For one token, only $K_t$ routed experts are evaluated. Counting
multiply--accumulate operations (MACs), the corresponding layer cost is
\[
  \operatorname{MACs}_t/\text{token}
  \approx
  3K_t\ell_t m_t
  +d\!\left(2\ell_t+N_t+3m_{s,t}\right)
  =3C_t+dQ_t.
\]
Under the convention that one MAC is two floating-point operations, the FLOP
count is twice this quantity. Matching $C_t$ and $Q_t$ therefore matches the
dominant activated computation. Activation functions, routing softmax,
weighted aggregation, and other lower-order elementwise operations are not
included in this approximation.

\paragraph{Expert-parallel communication.}
Figure~\ref{fig:communication-accounting} contrasts the communication paths.
In a standard MoE, routing one token to $K$ experts dispatches $K$ copies of a
$d$-dimensional representation to the workers hosting those experts. The
one-way logical payload is therefore $Kd$ scalars per token. The selected
expert outputs have the same width and produce another $Kd$ scalars on the
return all-to-all.

LatentMoE applies the down-projection on the source worker before dispatch.
Each expert assignment consequently sends an $\ell_t$-dimensional vector
rather than a $d$-dimensional one. Expert outputs return in the same latent
space and are aggregated before the up-projection. The shared projections
therefore remain outside the all-to-all, giving
\[
  V_t^{\text{one-way}}=K_t\ell_t
  \quad\text{and}\quad
  V_t^{\text{round-trip}}=2K_t\ell_t
  \qquad\text{scalars/token}.
\]
For $T$ routed tokens represented with $b$ bytes per scalar, these become
$bTK_t\ell_t$ bytes one way and $2bTK_t\ell_t$ bytes for dispatch plus return.
Routing requires $O(TK_t)$ indices and weights locally. Implementations that
transmit this metadata add an $O(TK_t)$ term, which is small relative to the
$O(TK_t\ell_t)$ activation payload at the widths studied here.

\subsection{System Throughput}
\label{app:system-throughput}

As shown in Table~\ref{tab:system-throughput}, we additionally compare
end-to-end training throughput. Our
heterogeneous allocation achieves throughput within measurement variation of
the uniform LatentMoE baseline, indicating that layerwise capacity
heterogeneity introduces negligible additional systems overhead. All
LatentMoE variants exhibit a modest throughput gap relative to the base model
under our current implementation, which does not use dedicated kernel
optimizations. This gap is consistent with implementation-level overhead from
finer-grained routing, smaller grouped GEMMs, and the latent projections.
Prior LatentMoE analysis shows that throughput-oriented large scale distributed MoEs can become strongly communication-bound and kernel-related overhead constitutes a smaller fraction of end-to-end execution time and may therefore be better amortized \citep{elango2026latentmoe}.
Thus, the heterogeneous
allocation has a near-homogeneous execution cost relative to uniform
LatentMoE, although LatentMoE itself retains a small wall-clock overhead
relative to the base model.

\begin{table}[H]
\centering
\small
\setlength{\tabcolsep}{6pt}
\caption{End-to-end training throughput.}
\label{tab:system-throughput}
\begin{tabular}{lc}
\toprule
Setting & Throughput (tokens/GPU/s) \\
\midrule
\multicolumn{2}{c}{\textbf{3.4B}} \\
\midrule
Base                      & 17991.2 \\
LatentMoE $\alpha=2$      & 17436.0 \\
LatentMoE $\alpha=4$      & 17319.2 \\
Ours                      & 17144.1 \\
\midrule
\multicolumn{2}{c}{\textbf{8B}} \\
\midrule
Base                      & 10139.6 \\
LatentMoE $\alpha=2$      & 9844.9 \\
LatentMoE $\alpha=4$      & 9810.8 \\
Ours                      & 9677.5 \\
\bottomrule
\end{tabular}
\end{table}

\section{Additional Expert-Shaping Results}

\subsection{Full Shape-Sweep Results}
\label{app:shape-sweep-results}

\begin{table}[H]
\centering
\small
\setlength{\tabcolsep}{5pt}
\caption{Complete configurations and validation losses for
Figure~\ref{fig:shape-sweeps}. $K/N$ denotes the number of activated experts
over the total number of experts.}
\label{tab:shape-sweep-results}
\begin{tabular}{rrrr@{\hspace{1.2cm}}rrrr}
\toprule
\multicolumn{4}{c}{8B-A560M} &
\multicolumn{4}{c}{3.4B-A250M} \\
$K/N$ & $\ell$ & $m$ & Val.\ loss &
$K/N$ & $\ell$ & $m$ & Val.\ loss \\
\midrule
$8/256$  & 320  & 2048 & 2.1151 & $8/256$  & 256  & 1536 & 2.2539 \\
$8/256$  & 512  & 1280 & 2.1064 & $8/256$  & 512  & 768  & 2.2385 \\
$8/256$  & 640  & 1024 & 2.1047 & $8/256$  & 640  & 616  & 2.2373 \\
$8/256$  & 816  & 800  & 2.1065 & $8/256$  & 768  & 512  & 2.2407 \\
$8/256$  & 1024 & 640  & 2.1074 & $8/256$  & 1024 & 384  & 2.2487 \\
$8/256$  & 1280 & 512  & 2.1105 & $16/512$ & 192  & 1024 & 2.2586 \\
$16/512$ & 160  & 2048 & 2.1423 & $16/512$ & 256  & 768  & 2.2407 \\
$16/512$ & 256  & 1280 & 2.1181 & $16/512$ & 320  & 616  & 2.2379 \\
$16/512$ & 320  & 1024 & 2.1108 & $16/512$ & 384  & 512  & 2.2320 \\
$16/512$ & 512  & 640  & 2.1028 & $16/512$ & 448  & 440  & 2.2281 \\
$16/512$ & 576  & 576  & 2.1024 & $16/512$ & 512  & 384  & 2.2299 \\
$16/512$ & 640  & 512  & 2.1038 & $16/512$ & 768  & 256  & 2.2419 \\
$16/512$ & 1024 & 320  & 2.1099 & $16/512$ & 1024 & 192  & 2.2462 \\
$16/512$ & 1280 & 256  & 2.1167 &          &      &      &        \\
\bottomrule
\end{tabular}
\end{table}

\subsection{Fixed Shared-Expert-Width Control}
\label{app:fixed-shared-control}

To remove shared-path reallocation as a confound, we repeat the
3.4B-A250M $(N,K)=(256,8)$ sweep with $m_s=768$ and the per-expert product
$\ell m\approx393{,}216$ fixed. All other training choices match the
corresponding shape sweep. Because the down- and up-projection parameter count
grows linearly with $\ell$, larger latent widths receive a modest parameter
advantage in this control. Table~\ref{tab:fixed-shared-control} reports this
advantage relative to $\ell=256$ as a fraction of the base model's total
trainable parameters, rather than as a fraction of activated parameters. The
largest advantage is only 0.47\%.

\begin{table}[H]
\centering
\small
\setlength{\tabcolsep}{9pt}
\caption{Validation loss in the fixed-$m_s$ control. The parameter advantage
is the additional down/up-projection parameters relative to $\ell=256$,
normalized by the total parameter count of the 3.4B-A250M base model.}
\label{tab:fixed-shared-control}
\begin{tabular}{rrrr}
\toprule
Latent width $\ell$ & Expert width $m$ & Val.\ loss
& Parameter advantage (\% total) \\
\midrule
256  & 1536 & 2.2517 & 0.00 \\
384  & 1024 & 2.2418 & 0.08 \\
512  & 768  & 2.2374 & 0.16 \\
640  & 616  & 2.2371 & 0.23 \\
768  & 512  & 2.2366 & 0.31 \\
1024 & 384  & 2.2456 & 0.47 \\
\bottomrule
\end{tabular}
\end{table}

The validation-loss span is 0.0151, compared with 0.0166 in the
parameter-matched sweep, and the broad near-square advantage is preserved.
Thus the shape effect remains even though the fixed-$m_s$ design increasingly
favors settings with larger latent widths.

\section{Additional Layerwise-Allocation Ablations}
\label{app:layerwise-ablation}

Table~\ref{tab:layerwise-ablation} reports the 3.4B sensitivity experiments
under the same training protocol as the main layerwise model. Allocations are
listed from shallow to deep, and each coefficient gives the number of MoE
blocks using that configuration.

\begin{table}[H]
\centering
\small
\setlength{\tabcolsep}{4pt}
\caption{Boundary and shape sensitivity at 3.4B. The first four rows use
the narrow and wide regimes from Table~\ref{tab:layerwise-configs}; the final
row is a matched-budget shape stress test. Lower loss is better.}
\label{tab:layerwise-ablation}
\begin{tabular}{lccr}
\toprule
Setting & $(\ell,m)$ & $(K,N)$ & Val.\ loss \\
\midrule
Three-block shallow
  & $3\times(352,448)+8\times(512,440)$
  & $3\times(20,640)+8\times(14,448)$
  & 2.2253 \\
ID-guided
  & $4\times(352,448)+7\times(512,440)$
  & $4\times(20,640)+7\times(14,448)$
  & $\mathbf{2.2234}$ \\
Five-block shallow
  & $5\times(352,448)+6\times(512,440)$
  & $5\times(20,640)+6\times(14,448)$
  & 2.2242 \\
Final block narrow
  & \begin{tabular}[t]{@{}c@{}}
    $5\times(352,448)+5\times(512,440)$\\$+\,1\times(352,448)$
    \end{tabular}
  & \begin{tabular}[t]{@{}c@{}}
    $5\times(20,640)+5\times(14,448)$\\$+\,1\times(20,640)$
    \end{tabular}
  & 2.2267 \\
\midrule
Strongly unbalanced shapes
  & $4\times(160,704)+7\times(880,224)$
  & $4\times(28,896)+7\times(16,512)$
  & 2.2498 \\
\bottomrule
\end{tabular}
\end{table}

Moving the boundary by one block changes validation loss by at most
$1.9\times10^{-3}$, whereas returning the final block to the narrow regime
incurs a larger penalty. The strongly unbalanced matched-budget control
performs substantially worse, consistent with the near-square design rule.

\section{Capacity Diagnostics}

\subsection{Realized Rank}
\label{app:realized-rank}

\newcommand{\RealizedRankRuns}{27}
\newcommand{\RealizedWeightExact}{1001}
\newcommand{\RealizedWeightTotal}{1059}
\newcommand{\RealizedWeightMin}{0.992}
\newcommand{\RealizedJacobianMin}{0.909}
\newcommand{\RealizedJacobianMax}{1.000}
\newcommand{\RealizedOutputMin}{0.988}
\newcommand{\RealizedOutputMax}{1.000}
\newcommand{\RealizedLatentMin}{0.995}
\newcommand{\RealizedLatentMax}{1.000}
\newcommand{\RealizedMixtureMin}{0.931}
\newcommand{\RealizedMixtureMax}{0.994}
\newcommand{\DepthPositiveRuns}{27}
\newcommand{\DepthTotalRuns}{27}
\newcommand{\DepthMedianRho}{+0.76}
\newcommand{\DepthPositiveWithoutFinal}{27}
\newcommand{\DepthMedianWithoutFinal}{+0.85}
\newcommand{\TransposePairs}{9}
\newcommand{\TransposeRankDiffMax}{3.4}
\newcommand{\PrecisionCases}{33}
\newcommand{\PrecisionExactCases}{33}
\newcommand{\RandomJacobianMin}{0.953}
\newcommand{\RandomJacobianMax}{1.000}

\paragraph{Estimator.}
For a SwiGLU expert, let
$\bm{a}=W_g\bm{z}$ and $\bm{b}=W_u\bm{z}$. We evaluate the exact local
Jacobian
\[
J_E(\bm{z}) =
W_o\!\left[
  \operatorname{diag}\!\big(\bm{b}\odot\phi'(\bm{a})\big)W_g
  +\operatorname{diag}\!\big(\phi(\bm{a})\big)W_u
\right]
\]
at tokens that were actually routed to the expert. The derivative includes
the implementation's activation-clamp masks, although no sampled activation
reaches the clamp. For singular values
$\sigma_1\geq\cdots\geq\sigma_s$, we estimate rank with the relative threshold
\[
  \operatorname{rank}_{\tau}(A)
  = \#\{j:\sigma_j(A)>\tau\sigma_1(A)\},
  \qquad \tau=10^{-3}.
\]
We apply the same estimator to each expert weight, to centered matrices of
expert inputs and outputs over routed tokens, and to the combined expert-map
Jacobian $\sum_{i\in T_K}p_iJ_{E_i}(\bm{z})$. The last quantity holds the
selected experts and routing weights locally fixed; it tests whether combining
the selected expert maps spans the latent space, rather than differentiating
through the router.

\paragraph{Sampling and controls.}
We collect 120,219 non-padding tokens from the same held-out data mixture used
for the ID study. In each MoE block we sample 16 experts with at least 512
routed tokens, evaluate 16 token-level Jacobians per expert, and retain up to
4,096 routed tokens per expert for the input and output clouds. The combined
expert-map statistic uses 24 tokens per block. Checkpoint weights are loaded
in their native bfloat16 representation and converted to float32 for the
SVD.

Two controls test the estimator. First, we construct matrices with known ranks
$0.25q$, $0.5q$, and $0.9q$, geometrically decaying nonzero spectra, and then
round them to bfloat16. The estimator recovers the true rank exactly in
\PrecisionExactCases/\PrecisionCases{} cases, ruling out a spurious
full-rank floor from bfloat16 quantization. Second, shape-matched iid Gaussian
experts attain Jacobian ranks between
\RandomJacobianMin$q$ and \RandomJacobianMax$q$. Their small relative deficit
near square shapes shows that the thresholded estimate there is affected by
the expected poorer conditioning, rather than implying unused trained
channels.

\paragraph{Full shape-sweep measurements.}
Tables~\ref{tab:realized-rank-3p4b} and
\ref{tab:realized-rank-8b} use exactly the configurations and row order of
Table~\ref{tab:shape-sweep-results}; validation loss is not repeated here.
$Y$ denotes the centered cloud of expert outputs, $Z$ the centered latent
inputs to a block, and $J_{\mathrm{mix}}$ the combined expert-map Jacobian
defined above. Every entry is a mean over MoE depth.

\begin{table}[H]
\centering
\scriptsize
\setlength{\tabcolsep}{3.0pt}
\caption{Realized ranks for the complete 3.4B-A250M shape sweep.
Weight and Jacobian columns are absolute ranks; the remaining columns are
normalized by their stated ceilings.}
\label{tab:realized-rank-3p4b}
\resizebox{\linewidth}{!}{%
\begin{tabular}{rrrrrrrrrrrr}
\toprule
$K/N$ & $\ell$ & $m$ & $q$ &
$W_g$ & $W_u$ & $W_o$ & $J$ & $J/q$ &
$Y/q$ & $Z/\ell$ & $J_{\mathrm{mix}}/\ell$ \\
\midrule

$8/256$ & 256 & 1536 & 256 & 256 & 256 & 256 & 253.6 & 0.991 & 1.000 & 1.000 & 0.994 \\
$8/256$ & 512 & 768 & 512 & 512 & 512 & 512 & 494.9 & 0.967 & 1.000 & 1.000 & 0.985 \\
$8/256$ & 640 & 616 & 616 & 616 & 616 & 616 & 566.8 & 0.920 & 0.999 & 1.000 & 0.980 \\
$8/256$ & 768 & 512 & 512 & 512 & 512 & 512 & 509.8 & 0.996 & 1.000 & 1.000 & 0.974 \\
$8/256$ & 1024 & 384 & 384 & 384 & 384 & 384 & 383.8 & 0.999 & 1.000 & 0.995 & 0.935 \\
$16/512$ & 192 & 1024 & 192 & 192 & 192 & 192 & 188.9 & 0.984 & 1.000 & 1.000 & 0.993 \\
$16/512$ & 256 & 768 & 256 & 256 & 256 & 256 & 251.5 & 0.983 & 1.000 & 1.000 & 0.992 \\
$16/512$ & 320 & 616 & 320 & 320 & 320 & 320 & 311.8 & 0.974 & 0.999 & 1.000 & 0.990 \\
$16/512$ & 384 & 512 & 384 & 384 & 384 & 384 & 368.7 & 0.960 & 1.000 & 1.000 & 0.990 \\
$16/512$ & 448 & 440 & 440 & 439.96 & 439.96 & 439.98 & 405.0 & 0.921 & 0.995 & 1.000 & 0.989 \\
$16/512$ & 512 & 384 & 384 & 384 & 384 & 384 & 381.2 & 0.993 & 1.000 & 1.000 & 0.985 \\
$16/512$ & 768 & 256 & 256 & 256 & 256 & 256 & 255.8 & 0.999 & 0.999 & 1.000 & 0.971 \\
$16/512$ & 1024 & 192 & 192 & 192 & 192 & 192 & 191.9 & 1.000 & 1.000 & 0.995 & 0.934

\tabularnewline
\bottomrule
\end{tabular}}
\end{table}

\begin{table}[H]
\centering
\scriptsize
\setlength{\tabcolsep}{3.0pt}
\caption{Realized ranks for the complete 8B-A560M shape sweep,
using the same definitions as Table~\ref{tab:realized-rank-3p4b}.}
\label{tab:realized-rank-8b}
\resizebox{\linewidth}{!}{%
\begin{tabular}{rrrrrrrrrrrr}
\toprule
$K/N$ & $\ell$ & $m$ & $q$ &
$W_g$ & $W_u$ & $W_o$ & $J$ & $J/q$ &
$Y/q$ & $Z/\ell$ & $J_{\mathrm{mix}}/\ell$ \\
\midrule

$8/256$ & 320 & 2048 & 320 & 320 & 320 & 320 & 316.7 & 0.990 & 1.000 & 1.000 & 0.993 \\
$8/256$ & 512 & 1280 & 512 & 512 & 512 & 512 & 502.7 & 0.982 & 1.000 & 1.000 & 0.990 \\
$8/256$ & 640 & 1024 & 640 & 640 & 640 & 640 & 619.0 & 0.967 & 0.999 & 1.000 & 0.987 \\
$8/256$ & 816 & 800 & 800 & 799.64 & 799.91 & 800 & 727.3 & 0.909 & 0.995 & 1.000 & 0.980 \\
$8/256$ & 1024 & 640 & 640 & 640 & 640 & 640 & 637.1 & 0.995 & 1.000 & 1.000 & 0.967 \\
$8/256$ & 1280 & 512 & 512 & 512 & 512 & 512 & 511.6 & 0.999 & 1.000 & 0.995 & 0.934 \\
$16/512$ & 160 & 2048 & 160 & 160 & 160 & 160 & 157.6 & 0.985 & 0.998 & 1.000 & 0.994 \\
$16/512$ & 256 & 1280 & 256 & 256 & 256 & 256 & 250.4 & 0.978 & 0.999 & 1.000 & 0.991 \\
$16/512$ & 320 & 1024 & 320 & 320 & 320 & 320 & 311.4 & 0.973 & 0.996 & 1.000 & 0.989 \\
$16/512$ & 512 & 640 & 512 & 512 & 512 & 512 & 486.5 & 0.950 & 0.999 & 1.000 & 0.988 \\
$16/512$ & 576 & 576 & 576 & 573.59 & 573.87 & 574.17 & 523.4 & 0.909 & 0.988 & 1.000 & 0.987 \\
$16/512$ & 640 & 512 & 512 & 512 & 512 & 512 & 503.8 & 0.984 & 1.000 & 1.000 & 0.985 \\
$16/512$ & 1024 & 320 & 320 & 320 & 320 & 320 & 319.8 & 1.000 & 1.000 & 1.000 & 0.964 \\
$16/512$ & 1280 & 256 & 256 & 256 & 256 & 256 & 256.0 & 1.000 & 1.000 & 0.995 & 0.931

\tabularnewline
\bottomrule
\end{tabular}}
\end{table}

\paragraph{The per-expert ceiling is attained.}
Across the \RealizedRankRuns{} sweep checkpoints,
\RealizedWeightExact{} of \RealizedWeightTotal{} block--matrix means equal
$q$ exactly, and the minimum is \RealizedWeightMin$q$. Local Jacobians attain
\RealizedJacobianMin--\RealizedJacobianMax$q$, while expert output clouds
attain \RealizedOutputMin--\RealizedOutputMax$q$. The latent inputs themselves
span \RealizedLatentMin--\RealizedLatentMax{} of their ambient width.
The full-rank weights and non-degenerate inputs make $q=\min(\ell,m)$ an
empirically relevant upper bound in trained models. Near-square experts
can show a slightly larger relative thresholding deficit, as they do in the
Gaussian control, but their larger $q$ still yields the largest absolute
realized ranks.

\begin{table}[H]
\centering
\small
\setlength{\tabcolsep}{4pt}
\caption{Ceiling--realized-rank relation within each controlled shape sweep.
$J$ is the depth-averaged realized local-Jacobian rank. The final column
counts all within-sweep pairs with distinct ceilings for which a larger $q$
also gives a larger $J$.}
\label{tab:ceiling-realized-rank}
\begin{tabular}{lcrrrrr}
\toprule
Scale & $K/N$ & Checkpoints & $r(q,J)$ & $\rho(q,J)$ & $J/q$ range & Ordered pairs \\
\midrule

3.4B & $8/256$ & 5 & 0.994 & 0.975 & 0.920--0.999 & 9/9 \\
3.4B & $16/512$ & 8 & 0.996 & 0.982 & 0.921--1.000 & 25/25 \\
8B & $8/256$ & 6 & 0.992 & 0.971 & 0.909--0.999 & 13/13 \\
8B & $16/512$ & 8 & 0.997 & 0.982 & 0.909--1.000 & 25/25

\tabularnewline
\bottomrule
\end{tabular}
\end{table}

Table~\ref{tab:ceiling-realized-rank} separates the four sweeps so that model
scale, routing configuration, training protocol, and approximate per-expert
parameter budget are fixed within each row. Across all 72 within-sweep pairs
with different ceilings, the configuration with the larger $q$ realizes the
larger absolute $J$. At the same time, the variation in $J/q$ confirms that
the ceiling and realized rank remain distinct. The high correlations show that
trained experts make use of an increased ceiling, while optimization and
validation loss retain additional dependencies.

\paragraph{Per-expert capacity versus mixture-level span.}
The combined expert-map statistic is already
\RealizedMixtureMin--\RealizedMixtureMax{} of $\ell$ in every configuration.
This is expected and does not contradict the per-expert capacity hypothesis.
Holding the selected experts and routing weights locally fixed,
\[
  \operatorname{rank}(J_{\mathrm{mix}})
  \leq
  \min\left(
    \ell,\sum_{i\in T_K}\operatorname{rank}(J_{E_i})
  \right)
  \leq \min(\ell,Kq).
\]
Every shape-sweep configuration satisfies $Kq\geq\ell$, so this
mixture-level bound is non-restrictive: the selected experts can collectively
span the latent space even though each constituent expert is individually
limited by $q$. Full mixture rank therefore certifies aggregate local span; it
does not imply that reducing the per-expert ceiling is lossless or that the
constituent conditional maps have equal capacity. The
$J_{\mathrm{mix}}$ statistic consequently rules out a mixture-level span
bottleneck, but it is not a test of the per-expert capacity hypothesis.

Across \TransposePairs{} within-sweep transposed pairs
$(\ell,m)$ and $(m,\ell)$, which share the same $q$, mean Jacobian rank differs
by at most \TransposeRankDiffMax\% of $q$, while
Table~\ref{tab:shape-sweep-results} shows different validation losses.
The rank proxy is therefore consistent with one aspect of local expert
capacity. Shape orientation, optimization, interactions with the shared path,
and mixture composition account for the remaining variation.

\paragraph{Interpretation for layerwise allocation.}
The allocation hypothesis concerns tolerance to per-expert compression rather
than numerical rank matching. We do not equate ID with $q_t$, nor do we
require a low-ID layer's mixture Jacobian to be rank deficient. Instead, the
directional hypothesis is that reducing the per-expert ceiling incurs a
smaller loss penalty in lower-ID regions than in higher-ID regions. The
matched-count exchange preserves the same multiset of narrow and wide block
configurations while changing their placement across depth, providing a
controlled test of this directional claim rather than a calibrated mapping
from ID values to rank ceilings.

\paragraph{Depth-wise evidence and the final-block exception.}
Uniform-shape models keep $q$ fixed across depth, so any layerwise variation
in realized rank is not induced by a changing rank budget. Correlating each
checkpoint's block-averaged local-Jacobian rank with the corresponding
base-model TwoNN profile gives a positive Spearman correlation in
\DepthPositiveRuns/\DepthTotalRuns{} checkpoints, with median
$\rho=\DepthMedianRho$. Numerical rank is already near saturation, so the
variation is small; this is directional rather than magnitude evidence.
The final MoE block frequently departs from the preceding trend. Reporting
the prespecified sensitivity analysis rather than silently dropping it, the
correlation is positive in \DepthPositiveWithoutFinal/\DepthTotalRuns{}
checkpoints with median $\rho=\DepthMedianWithoutFinal$ when that block is
excluded. The boundary ablation in Section~\ref{sec:pretraining-results}
nevertheless favors the higher-capacity configuration for the final block,
consistent with its ID declining only mildly and remaining well above the
low-ID shallow prefix (Table~\ref{tab:id-cross-scale-values}).

Finally, the two-regime models produce the intended absolute-rank step while
retaining the matched routed budgets. Table~\ref{tab:layerwise-realized-rank}
shows that the deep regime realizes a larger local Jacobian rank at both
scales.

\begin{table}[H]
\centering
\small
\setlength{\tabcolsep}{5pt}
\caption{Realized rank in the shallow and deep regimes of the two-segment
models. ID and rank are averaged within each segment.}
\label{tab:layerwise-realized-rank}
\begin{tabular}{lrrrrrr}
\toprule
Scale & Blocks & $\ell\times m$ & $q$ & Mean ID & $J$ & $J/q$ \\
\midrule

3.4B & 1--4 & $352\!\times\!448$ & 352 & 4.78 & 334.3 & 0.950 \\
3.4B & 5--11 & $512\!\times\!440$ & 440 & 12.97 & 431.2 & 0.980 \\
8B & 1--5 & $464\!\times\!512$ & 464 & 4.87 & 425.0 & 0.916 \\
8B & 6--15 & $640\!\times\!512$ & 512 & 13.07 & 504.9 & 0.986

\tabularnewline
\bottomrule
\end{tabular}
\end{table}

\begin{figure}[htbp]
\centering
\includegraphics[width=\linewidth]{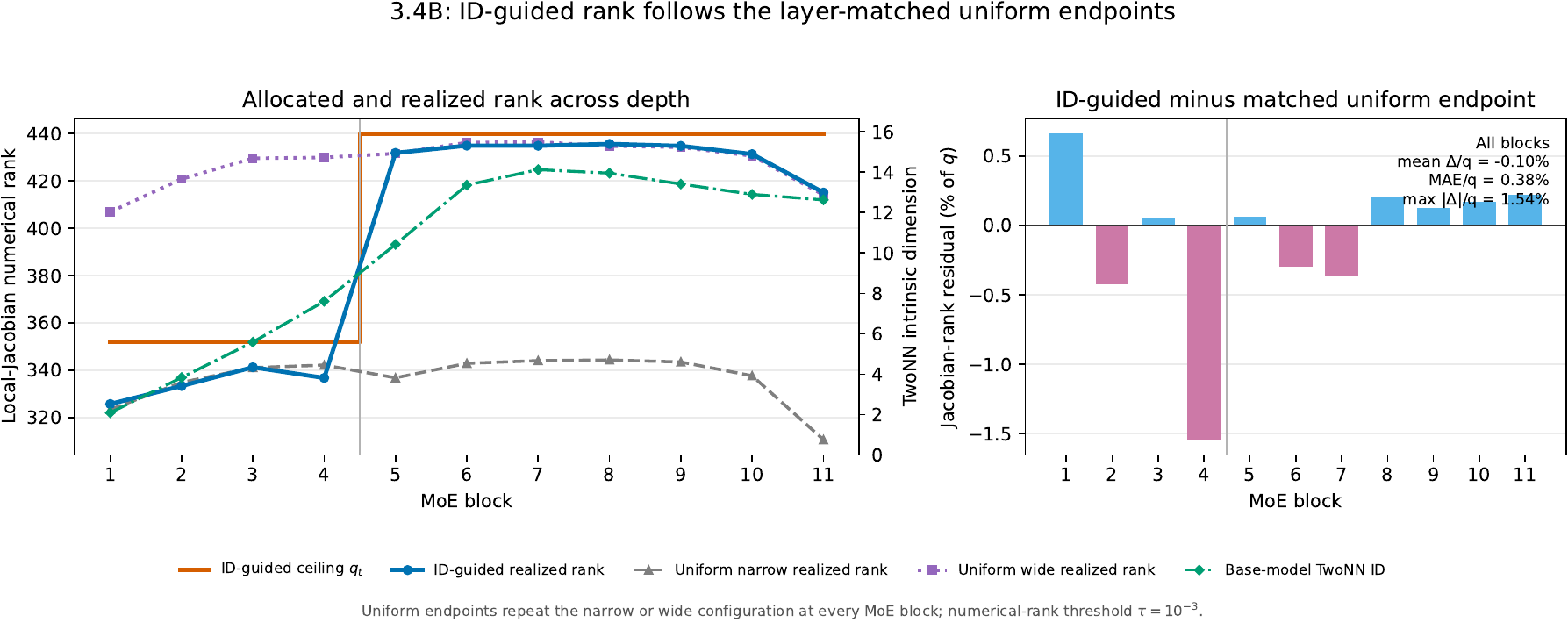}
\caption{Matched-endpoint realized-rank check at 3.4B. \textbf{Left:} the
ID-guided checkpoint's blockwise local-Jacobian numerical rank, its allocated
ceiling $q_t$, and the realized ranks of the two uniform endpoint checkpoints.
The uniform narrow and wide checkpoints repeat their respective
$(\ell,m,K,N)$ configurations at every MoE block. Base-model TwoNN ID is
shown only as the ordinal signal used to choose the allocation; its numerical
scale is not comparable to rank. \textbf{Right:} the ID-guided rank minus the
layer-matched uniform endpoint, normalized by $q_t$.}
\label{fig:layerwise-matched-rank}
\end{figure}

\paragraph{Matched-endpoint check.}
The segment averages above verify that the allocated rank changes across the
regime boundary. To separate this designed change in $q_t$ from a possible
effect of heterogeneous training, we additionally compare the 3.4B
two-segment checkpoint at every block with the uniform checkpoint that repeats
the same $(\ell,m,K,N)$ configuration at all blocks. Thus, the matched
counterfactual is the uniform narrow model in blocks 1--4 and the uniform wide
model in blocks 5--11. Figure~\ref{fig:layerwise-matched-rank} shows that the
ID-guided checkpoint closely follows this piecewise uniform reference: the
mean signed residual is $-0.10\%$ of $q$, the mean absolute residual is
$0.38\%$ of $q$, and the maximum absolute residual is $1.54\%$ of $q$.
This is a mechanism check that the heterogeneous model realizes its allocated
ceiling similarly to the corresponding uniform models.

\subsection{Intrinsic-Dimension Estimation}
\label{app:id-details}

At each block boundary, we collect 20,000 non-padding token representations
from the same held-out corpus and randomly retain 5,000 for TwoNN estimation.
Both scales use a maximum sequence length of 2,048, seed 0, and mean reduction
over the model's parallel chain dimension. We exclude the embedding output
from regime selection because repeated discrete token vectors do not satisfy
TwoNN's local continuous-manifold assumption.

The cross-scale comparison uses the final 3.4B and 8B base checkpoints.
Table~\ref{tab:id-cross-scale-values} gives the complete layerwise values,
including the embedding output $L0$ for reference.

\begin{table}[H]
\centering
\small
\setlength{\tabcolsep}{5pt}
\caption{Layerwise TwoNN intrinsic-dimension values underlying the left panel
of Figure~\ref{fig:id-profiles}. Relative depth is $t/L$, where $L$ is the
number of Transformer blocks; $L0$ denotes the embedding output.}
\label{tab:id-cross-scale-values}
\begin{tabular}{rrr@{\hspace{1.0cm}}rrr}
\toprule
\multicolumn{3}{c}{3.4B-A250M ($L=12$)} &
\multicolumn{3}{c}{8B-A560M ($L=16$)} \\
Layer & $t/L$ & ID & Layer & $t/L$ & ID \\
\midrule

L0 & 0.000 & 0.194 & L0 & 0.000 & 0.183 \\
L1 & 0.083 & 2.092 & L1 & 0.062 & 2.081 \\
L2 & 0.167 & 3.835 & L2 & 0.125 & 3.631 \\
L3 & 0.250 & 5.587 & L3 & 0.188 & 4.839 \\
L4 & 0.333 & 7.607 & L4 & 0.250 & 5.970 \\
L5 & 0.417 & 10.423 & L5 & 0.312 & 7.833 \\
L6 & 0.500 & 13.360 & L6 & 0.375 & 10.452 \\
L7 & 0.583 & 14.125 & L7 & 0.438 & 12.781 \\
L8 & 0.667 & 13.951 & L8 & 0.500 & 13.967 \\
L9 & 0.750 & 13.413 & L9 & 0.562 & 15.002 \\
L10 & 0.833 & 12.901 & L10 & 0.625 & 14.854 \\
L11 & 0.917 & 12.627 & L11 & 0.688 & 13.979 \\
L12 & 1.000 & 9.987 & L12 & 0.750 & 13.371 \\
-- & -- & -- & L13 & 0.812 & 12.902 \\
-- & -- & -- & L14 & 0.875 & 12.092 \\
-- & -- & -- & L15 & 0.938 & 11.315 \\
-- & -- & -- & L16 & 1.000 & 10.446 \\

\tabularnewline
\bottomrule
\end{tabular}
\end{table}

To measure evolution during training, we use a separate 8B checkpoint series
at 2k, 4k, 10k, 16k, and 20k steps under the same estimation protocol.
Table~\ref{tab:id-training-values} reports every layer and checkpoint.

\begin{table}[H]
\centering
\small
\setlength{\tabcolsep}{7pt}
\caption{Layerwise TwoNN intrinsic dimension over training for the separate
8B checkpoint series underlying the right panel of
Figure~\ref{fig:id-profiles}.}
\label{tab:id-training-values}
\begin{tabular}{rrrrrr}
\toprule
Layer & 2k & 4k & 10k & 16k & 20k \\
\midrule

L0 & 0.161 & 0.183 & 0.189 & 0.192 & 0.192 \\
L1 & 1.761 & 1.753 & 1.921 & 2.094 & 2.163 \\
L2 & 2.757 & 3.085 & 3.413 & 3.637 & 3.705 \\
L3 & 4.137 & 4.265 & 4.755 & 4.993 & 5.144 \\
L4 & 5.061 & 4.884 & 5.246 & 5.571 & 5.614 \\
L5 & 6.235 & 6.059 & 6.505 & 6.873 & 6.928 \\
L6 & 7.512 & 7.979 & 8.800 & 9.151 & 8.992 \\
L7 & 9.440 & 10.425 & 11.618 & 11.832 & 11.604 \\
L8 & 10.882 & 11.602 & 12.297 & 11.823 & 11.433 \\
L9 & 11.769 & 12.728 & 12.805 & 12.370 & 12.036 \\
L10 & 12.785 & 14.327 & 13.975 & 13.302 & 12.778 \\
L11 & 12.976 & 13.903 & 13.920 & 13.125 & 12.449 \\
L12 & 12.865 & 13.712 & 13.685 & 12.939 & 12.316 \\
L13 & 12.847 & 13.437 & 12.997 & 12.255 & 11.681 \\
L14 & 12.067 & 12.451 & 12.147 & 11.568 & 10.964 \\
L15 & 11.376 & 11.750 & 11.518 & 10.967 & 10.356 \\
L16 & 10.662 & 10.110 & 9.777 & 9.515 & 9.188 \\

\tabularnewline
\bottomrule
\end{tabular}
\end{table}

Across the two model sizes and all five 8B checkpoints, the depth-wise trends
are close: ID rises through the early blocks, peaks in the middle-to-deep
region, and declines toward the output.
\end{document}